\documentclass[runningheads]{llncs}

\usepackage{graphicx}
\usepackage[dvipsnames]{xcolor}
\usepackage{amsmath, amssymb, amsfonts}
\usepackage{xspace}

\usepackage{lineno}

\newcommand{\AM}[1]{}
\newcommand{\LA}[1]{}
\newcommand{\MT}[1]{}
\newcommand{\todo}[1]{}

\makeatletter
\DeclareRobustCommand\onedot{\futurelet\@let@token\@onedot}
\def\@onedot{\ifx\@let@token.\else.\null\fi\xspace}

\def\eg{\emph{e.g}\onedot} 
\def\ie{\emph{i.e}\onedot}

\newcommand{\printfnsymbol}[1]{%
  \textsuperscript{\@fnsymbol{#1}}%
}

\newcommand{\bR}{\ensuremath \mathbb{R}}

\makeatother

\begin{document}

\title{Learning to Approximate Directional Fields Defined over 2D Planes}


\author{Maria Taktasheva\thanks{These two authors contribute equally to the work.} \and
Albert Matveev\printfnsymbol{1} \and
Alexey Artemov \and
Evgeny Burnaev}

\authorrunning{M. Taktasheva et al.}

\institute{Skolkovo Institute of Science and Technology, Moscow, Russia}

\maketitle          

\begin{abstract}
Reconstruction of directional fields is a need in many geometry processing tasks, such as image tracing, extraction of 3D geometric features, and finding principal surface directions. A common approach to the construction of directional fields from data relies on complex optimization procedures, which are usually poorly formalizable, require a considerable computational effort, and do not transfer across applications. In this work, we propose a deep learning-based approach and study the expressive power and generalization ability.

\end{abstract}

\section{Introduction}
\label{sec:intro}

Many spatially varying geometric and physical properties of objects, such as surface curvature, stress tensors, or gradients of scalar fields, are commonly described by directional \emph{(vector)} fields: certain quantities (generally vector-valued) assigned to respective points in some spatial domain. 
While real-world measurements correspond to our most intuitive notion of directional fields, they can just as easily be \emph{synthesized} via optimization by computational models accounting for constraints such as physical realizations or alignment conditions. Corresponding to such requirements, different representations for directional fields have been proposed, differing, \eg, by several directional entities per point of the domain, or symmetries between them. 

In the area of 2D and 3D computer graphics and geometry processing, directional fields have been utilized for a vast number of applications, such as mesh generation using distinct 3D data modalities~\cite{bommes2013quad,diamanti2014designing}, texture mapping~\cite{ray2010invisible}, and image-based tracing of line drawings~\cite{bessmeltsev2019vectorization}, to name a few. Appearing datasets~\cite{koch2019abc} open more possibilities for this setup with various problems to solve. With the guidance of an appropriately designed directional field, both topological (\eg, placement of singularity points) and geometric (\eg, smoothness) properties of the underlying geometric structure may be efficiently derived. Other applications which could benefit from learnable directional fields include remote sensing~\cite{kolos2019procedural,bokhovkin2019boundary,RSdamage2018,ChangeDetectionRS2019}, RGBD data processing~\cite{voinov2018perceptually} and related applications \cite{keypoint,osin}, shape retrieval~\cite{notchenko2017large}.

However, obtaining a robust approximation of a directional field from raw input data is a challenging problem in many instances. Current approaches to computing directional fields require optimization of non-trivial targets with custom optimizers (\eg, ADMM, L-BFGS, and their versions), which may be computationally demanding and may yield unstable solutions, where significant noise, occlusions, or gaps exist in the data~\cite{bessmeltsev2019vectorization}.

On the other hand, modern deep convolutional neural networks (CNNs) have proven themselves effective in learning arbitrarily complex functions from both theoretical and practical standpoints~\cite{hornik1989multilayer,mhaskar2016deep}. 
However, little work has been done to approximate directional fields using learning-based approaches; thus, questions exist of whether training a conventional CNN to approximate a directional field would be easy or even feasible, and whether it can learn to produce highly robust directional fields.

In this work, we conduct an initial feasibility study, aiming to establish a principled learning setup for an approximation of directional fields. 
For our study, we take a setting simplified in several aspects. First, we restrict ourselves to directional fields defined over subsets of 2D planes, \ie over rectangular regions $\Omega \subset \bR^2$, enabling us to use conventional CNN architectures. Moreover, in $\bR^2$, an explicit field representation using Euclidean coordinates is straightforward, which is not the case on, \eg, curved surfaces~\cite{vaxman2016directional}. Next, without loss of generality, we focus on 2-PolyVector fields~\cite{diamanti2014designing} as a structure of our probe directional field. Lastly, to obtain a particular ground-truth for our experiments, we opt to derive the field from geometric primitives encountered in vector computer graphics, as they allow for constructing an explicit analytical description of the result. The overall setup represents a conceptually simple but important instance, as it corresponds to several practical applications, such as image tracing~\cite{bessmeltsev2019vectorization}. We carry out two experiments, aiming to (1) establish the effectiveness of a deep CNN in an approximation setting, and (2) evaluate its generalization ability.

This short paper is organized as follows. In Section~\ref{sec:design}, we detail the design of our probe directional field, our choice of the learning algorithm, and the formation of our training data. Section~\ref{sec:experiments} presents two experiments evaluating our architecture. We conclude in Section~\ref{sec:conclusion} with a brief discussion. 

\begin{figure}[t]
\begin{center}
\includegraphics[width=\columnwidth]{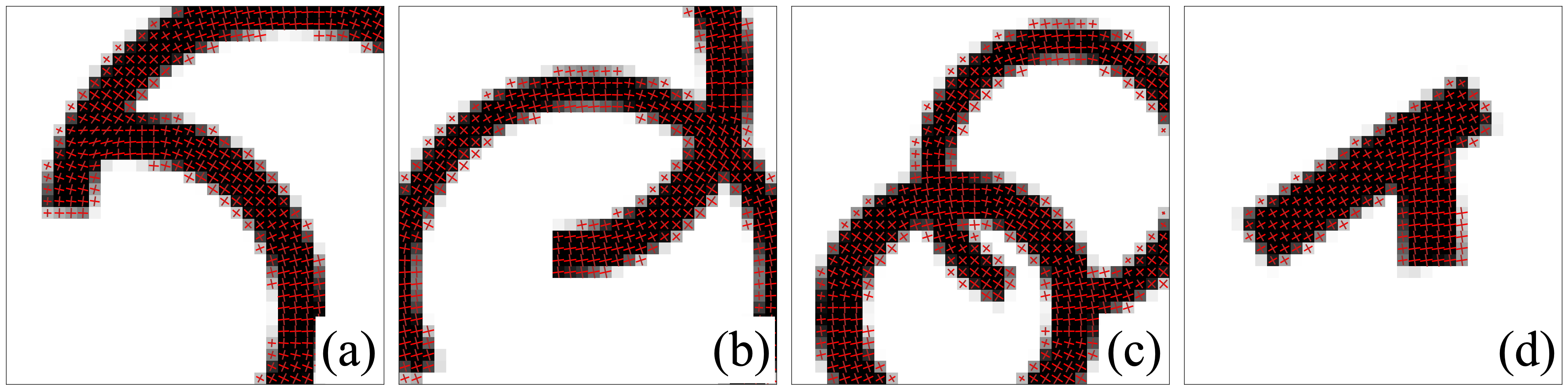}
\caption{(a)--(d): examples of rasterized vector primitives with accompanying discretized ground-truth 2-PolyVector field derived according to our scheme (see text).}
\label{fig:ground-truth-examples}
\end{center}
\end{figure}

\section{Directional field design and the approximation setup}
\label{sec:design}

\subsection{Our probe directional field}
For the sake of simplicity, we define our directional fields over gray-scale images. Specifically, we assign two complex-valued vectors $\{u, v\}$ to each pixel $x \in \Omega$ which is above the given threshold\LA{clarify}. Following the established nomenclature~\cite{vaxman2016directional,de2016vector}, we denote our directional fields as \emph{2-PolyVector fields}.
The field is obtained through an optimization procedure on a complex polynomial~\cite{vaxman2016directional} 
\[
    f(z) = (z^2 - u^2)(z^2 - v^2) = z^4 + c_2 z^2 + c_0,
\]
where the coefficients are indexed by the corresponding power of the variable.

While directional fields such as ours can be defined over arbitrary images, we define them on a particular class of images: line drawings, \ie, collections of 1D geometric primitives such as straight and curved lines. For points on curves, the first field component is aligned with the curve tangent; the second field component is an interpolation\LA{clarify} between the curve normal and the tangent of a nearest intersecting curve. For points not on curves, we leave the field undefined. Thus, we obtain an intuitive interpretation of field components $\{u, v\}$; the PolyVector structure allows us to express the directionality of curves. A similar parameterization is adopted in~\cite{bessmeltsev2019vectorization}.

Leveraging the analytic expressions for the primitives constituting the line drawing image (\eg, lines, arcs, and B{\'e}zier curves), we can define the first field vector as a unit tangent vector of a curve. The unit tangent is uniquely determined by its slope angle $\alpha$, with its complex coordinates $u = (\cos \alpha, \sin \alpha)$. The second vector $v = (\cos \beta, \sin \beta)$ is also determined by its slope angle $\beta$; $v$ can be obtained based on the following simple rules:
\begin{itemize}
    \item if two curves intersect in one pixel, as a tangent vector of a second curve;
    \item if the intersection is relatively far, as a normal vector of a curve;
    \item if the intersection is not far --- as an interpolation between these two cases\LA{which one do we choose?}.
\end{itemize}
We assume that junctions of three and more are rare and only consider junctions of two curves in our model. The definition of a PolyVector field is straightforward to extend to capture other types of junctions.

To become independent to the order of primitives and the chosen sign of a tangent, we follow a well-known workaround and perform variable substitution $c_0 = u^2 v^2, c_2 = -(u^2 + v^2),$
which determines two vectors $u$ and $v$ up to relabeling and sign~\cite{bessmeltsev2019vectorization}. The final vector field representation has four parameters: two complex coordinates for each of the variables $c_0$ and $c_2$. 
We interpolate between the field vectors by smoothing a field within channels related to the same complex vector with Gaussian kernel\LA{what?}. 
We display typical raster line-drawing images along with the corresponding ground-truth directional fields in Fig.~\ref{fig:ground-truth-examples}.


\subsection{Our neural network-based direction field approximator}

\noindent\textbf{Architecture. }
We formulate our field approximation as a multivariate regression task, where one needs to obtain an output of the same shape as the input, except for the number of output channels. When designing our directional field approximation architecture, we take inspiration from recent progress in semantic segmentation and derive our network from the vanilla U-Net model~\cite{ronneberger2015u}. 

U-Net architecture consists of the encoder and decoder parts with skip connections between them. Each encoder layer has two convolutional layers with the increasing number of channels, followed by batch normalization and ReLU activation function. Each encoder layer output is fed through max pooling. The decoder has a symmetric structure, where the numbers of channels of convolutional layers decrease, and the upsampling layer is put instead of max pooling. The output of each of the encoder layers is concatenated with the input of the corresponding decoder layer so that the information is passed through skip connections. We modify the architecture by replacing the softmax layer at the end of the network with a linear layer having four output channels, two for real parts of $c_0,\, c_2$, two for imaginary parts.

We implemented the network in pytorch~\cite{paszke2017automatic} and trained it on one NVIDIA Tesla P100 with 16Gb GPU memory. The training process took 15 minutes for single image experiments and 40 minutes for the generalization study.

\noindent\textbf{Loss function. }
We base our loss function on the results from~\cite{bessmeltsev2019vectorization}, where the PolyVector field is found as a solution of a complex optimization problem and $c_0, c_2$ are treated as complex functions defined at each pixel. The optimizer searches for a solution which is aligned with a tangent field approximation obtained by applying a Sobel filter. As an alignment term, we use mean squared error loss for real and imaginary parts of complex coefficients $c_0$ and $c_2$. The smoothness of the resulting field is guaranteed by a smoothness regularization term, formulated as the difference in \(c_k,\, k = 0, 2\) between the neighbouring pixels (if these are available), where \( \nabla c_k(x_{i, j}) \) is a complex number:
\begin{align*}
    \int \| \nabla c_k(x) \|^2 dx &= \sum_{i} \sum_{j} \| \nabla c_k(x_{i,j}) \|^2 \\
    &=  \sum_{i} \sum_{j}  Re(\nabla c_k(x_{i, j}))^2 + Im(\nabla c_k(x_{i, j}))^2, \\
    \nabla c_k(x_{i, j}) &= \text{vertical change} + \text{horizontal change},\ k=0,2
\end{align*}
where the real and imaginary part are represented as separate channels.

The overall loss we optimize is the following:
\begin{equation*}
    L(c_0,c_2,c_0^*,c_2^*) = \sum_{k=0,2} \sum_i \sum_j \|c_k(x_{i,j}) - c_k^*(x_{i,j})\|^2 + \gamma \sum_{k=0,2} \sum_{i} \sum_{j} \| \nabla c_k(x_{i,j}) \|^2
\end{equation*}



\begin{figure}[h!]
\begin{center}
\includegraphics[width=0.8\columnwidth]{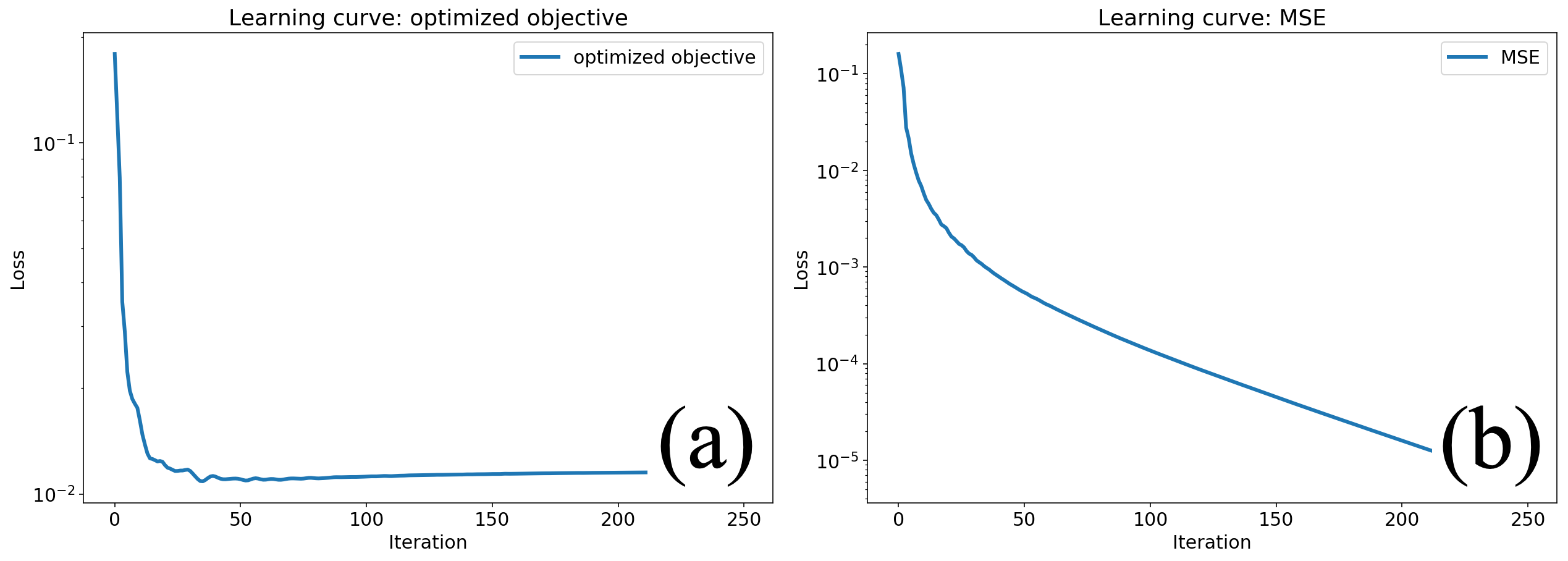}
\caption{Approximation progress, single image: (a) optimized objective, (b) MSE loss.}
\label{fig:pvf-expressive-learning-curves}
\end{center}
\end{figure}

\begin{figure}[h]
\begin{center}
\vspace*{-.5cm}
\includegraphics[width=\columnwidth]{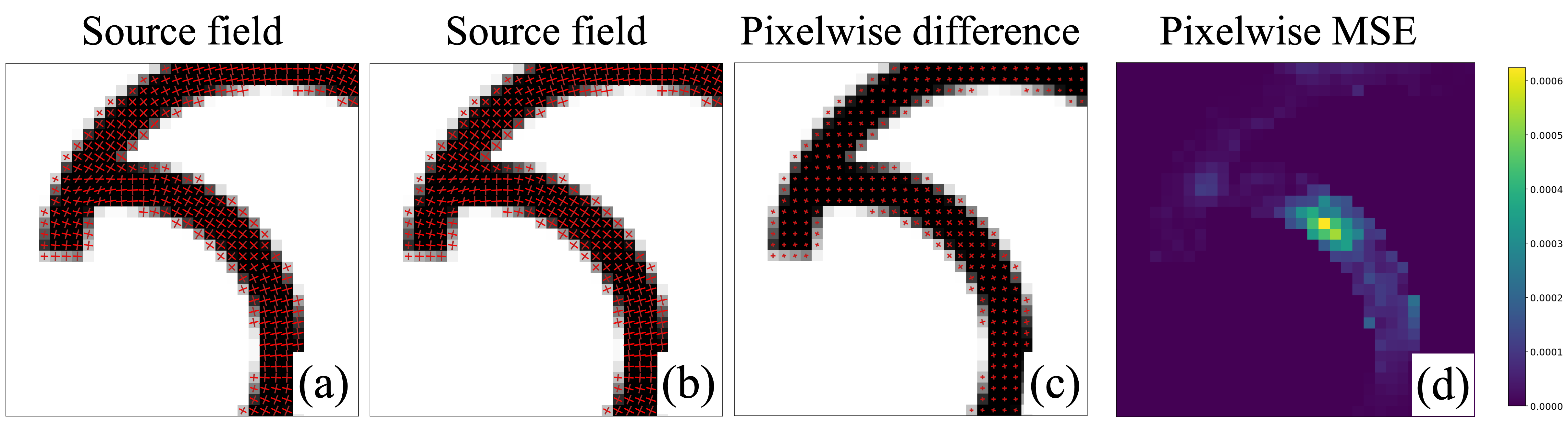}
\caption{Approximation results, single image: (a) input drawing and ground truth field, (b) approximated directional field, (c) difference between ground truth and approximated fields, (d) error heatmap. Best viewed in zoom.}
\label{fig:pvf-expressive-raster}
\end{center}
\end{figure}

\section{Experimental results}
\label{sec:experiments}

\subsection{Expressive power of a conventional CNN}
Here we show that a common CNN architecture optimized with stochastic gradient descent is capable of representing a directional field. We select a random synthetic $64 \times 64$ image with two primitives and optimize network parameters for 50 iterations. We display learning curves in Fig.~\ref{fig:pvf-expressive-learning-curves} and the resulting approximations in Fig.~\ref{fig:pvf-expressive-raster}. We conclude that our network learns to represent the input directional field with an alignment error of $10^{-5}$ order of magnitude and regularized loss of $10^{-2}$ order of magnitude.

\subsection{Generalization study}
\begin{figure}[h!]
\begin{center}
\includegraphics[width=0.8\columnwidth]{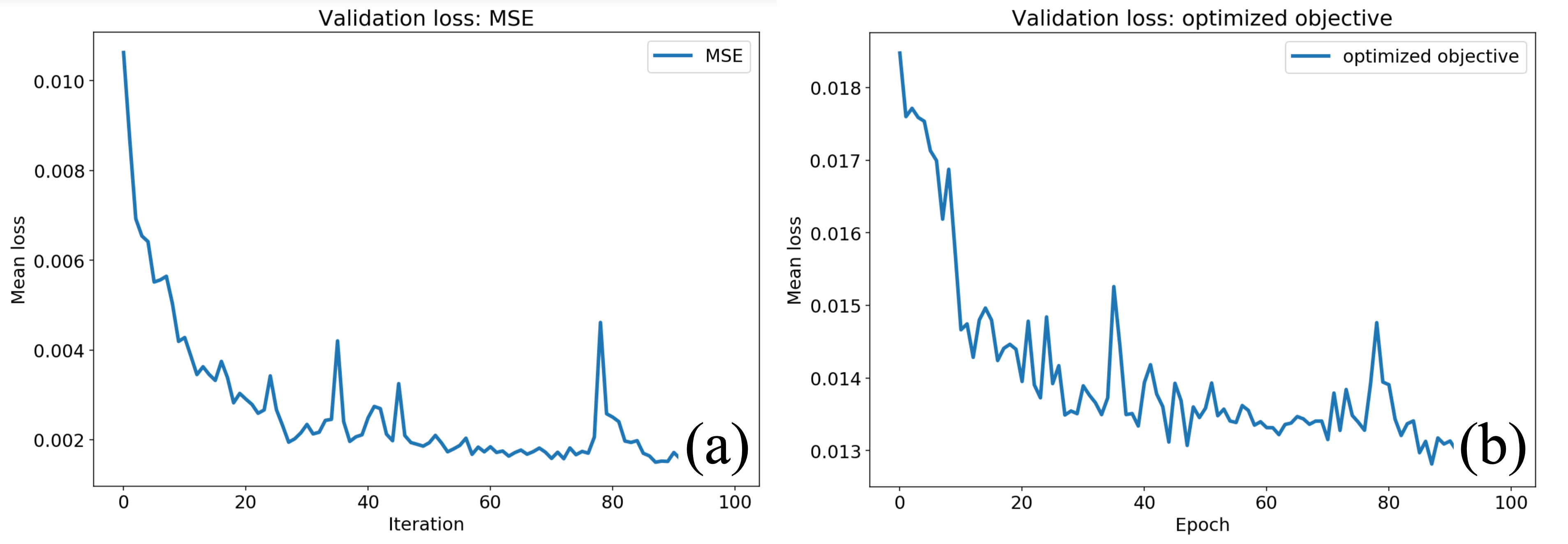}
\caption{Learning progress, 5000 images: (a) MSE loss, (b) optimized objective.}
\label{fig:pvf-generalization-learning-curves}
\end{center}
\end{figure}

\begin{figure}[h!]
\begin{center}
\vspace*{-.5cm}
\includegraphics[width=\columnwidth]{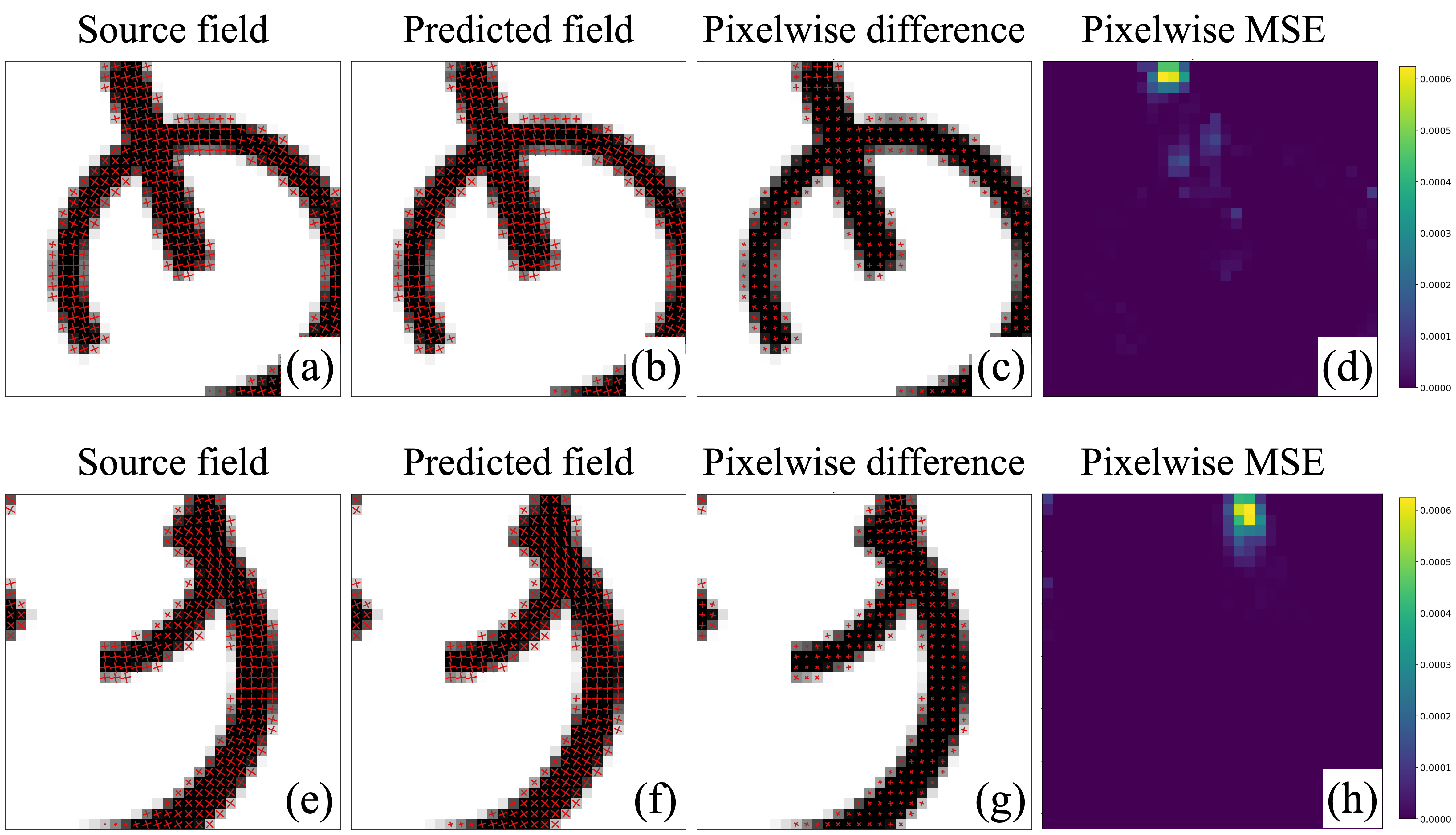}
\caption{Learning progress, 5000 images: (a),(e) input drawing and ground truth field, (b),(f) approximated directional field, (c),(g) difference between ground truth and approximated fields, (d),(h) error heatmap. Best viewed in zoom.}
\label{fig:pvf-generalization-raster}
\end{center}
\end{figure}

The second experiment is a proof of concept to the learnability of a directional field on a dataset of sufficient size. We synthesize and rasterize 5500 $64 \times 64$-pixel line drawings, splitting them into 5000 training and 500 validation images. We train the network using our data for 100 epochs. In Fig.~\ref{fig:pvf-generalization-learning-curves}, we plot learning curves, showing the resulting predictions on validation samples in Fig.~\ref{fig:pvf-generalization-raster}. We conclude that our model can learn the underlying directional field for the generated dataset: we report validation MSE loss $0.00152$ and regularized loss $0.01321$.

\section{Conclusion}
\label{sec:conclusion}

In this work, we have established two simple but important facts that (1) a general PolyVector field can be efficiently represented using an off-the-shelf CNN, and (2) the same CNN can generalize to unseen instances by training on a synthetic dataset of line drawings. These findings strongly motivate the need for further research on learnable methods for directional fields approximation and processing. 

\section*{Acknowlegement}

The work was supported by the Russian Science Foundation under Grant 19-41-04109.

%
%
\bibliographystyle{splncs04}
\bibliography{00-main.bib}
%





\end{document}